%% file: main.tex
\documentclass{article}

\PassOptionsToPackage{numbers, sort, compress}{natbib}

\usepackage[preprint]{neurips_2026}

\usepackage[utf8]{inputenc} 
\usepackage[T1]{fontenc}    
\usepackage{hyperref}       
\usepackage{url}            
\usepackage{booktabs}       
\usepackage{amsfonts}       
\usepackage{nicefrac}       
\usepackage{microtype}      
\usepackage{xcolor}         

\title{Deep Arguing}

\author{%
  Adam Gould\thanks{Corresponding Author} \\
  Department of Computing\\
  Imperial College London\\
  \texttt{adam.gould19@imperial.ac.uk} \\
  \And
  Francesca Toni \\
  Department of Computing\\
  Imperial College London \\
  \texttt{f.toni@imperial.ac.uk} \\
}

\input{header}

\begin{document}





\maketitle

\begin{abstract}

Deep learning has become the dominant approach for creating high capacity, scalable models across diverse data modalities. 
However, because these models rely on a large number of learned parameters, tightly couple feature extraction with task objectives, and often lack explicit reasoning mechanisms, it is difficult for humans to understand how they arrive at their predictions.
Understanding what representations emerge and why they arise 
from the training data remains an open challenge.
We introduce \textit{Deep Arguing}, a novel neurosymbolic approach that integrates deep learning with argumentation construction and reasoning for interpretable classification with different data modalities. In 
our approach deep neural networks construct an argumentation structure wherein data points support their assigned label 
and attack  
different ones. Using differentiable argumentation semantics for reasoning, the model is trained end-to-end to jointly learn feature representation and argumentative interactions. This results in argumentation structures providing faithful case-based explanations for predictions. Structure constraints over the argumentation graph guide learning, improving both interpretability and  predictive performance. Experiments with tabular and imaging datasets show that Deep Arguing achieves performance competitive with standard baselines whilst offering interpretable argumentative reasoning.

\end{abstract}

\section{Introduction}
\label{sec:intro}

\input{introduction/introduction}

\section{Related work}
\label{sec:related-work}
\input{introduction/related}

\section{Preliminaries}
\label{sec:preliminaries}
\input{preliminaries/preliminaries}

\section{Deep Arguing}
\label{sec:methodology}

\input{methodology/methodology}

\section{Experiments}
\label{sec:experiments}

\input{experiments/experiments}

\section{Deep Arguing interpretability}
\label{sec:interpretability}
\input{interpretability/interpretability}

\section{Conclusion}
\label{sec:conclusion}
\input{conclusion/conclusion}


\todoin{
\begin{itemize}
    \itemtodo Verify styling requirements
    \itemtodo Verify all references
    \itemtodo Ensure references are numbered correctly
    \itemtodo Complete the checklist
    \itemtodo FT: do a final check at some stage: argumentaiton rather than argumentative/argument graph allover?
    \itemtodo Number the eqns
\end{itemize}
}

\begin{ack}
Research by Adam Gould was supported by UK Research and Innovation [UKRI Centre for Doctoral Training in AI for Healthcare grant number EP/S023283/1].
Francesca Toni was partially funded by the ERC under the EU’s Horizon 2020 research and innovation programme (grant agreement No. 101020934). Toni was also funded by EPSRC (grant UKRI3928, NeSyDebates).
 
\end{ack}


\bibliographystyle{plain}
\bibliography{references}


\appendix

\section{Minimality condition}
\input{appendix/minimality}

\section{Training algorithm}
\input{appendix/training-algorithm}

\section{Experiment details}
\input{appendix/experiment-details}

\section{Experiments Discussion}
\input{appendix/experiment-discussion}

\section{Full QBAF}
\input{appendix/full-qbaf}



\end{document}

%% file: header.tex
\usepackage{ amsmath }
\usepackage{ amssymb }
\usepackage{amsfonts}

\usepackage{enumitem}
\usepackage{algorithm}
\usepackage{algpseudocode}

\usepackage{tikz}
\usetikzlibrary{decorations.pathmorphing,shapes,positioning,arrows.meta,arrows,automata,positioning,calc,shadows}
\usepackage{caption}
\usepackage{subcaption}
\usepackage{graphicx}

\usepackage{xargs} 
\let\todo\relax
\usepackage[disable]{todonotes}
\setuptodonotes{size=tiny}

\newcommandx{\complete}[2][1=]{\todo[linecolor=orange,backgroundcolor=orange!25,bordercolor=orange,#1]{Complete: #2}}
\newcommandx{\tocite}[2][1=]{\todo[linecolor=pink,backgroundcolor=pink!25,bordercolor=pink,#1]{Cite: #2}}
\newcommandx{\unsure}[2][1=]{\todo[linecolor=blue,backgroundcolor=blue!5,bordercolor=blue,#1]{Unsure: #2}}
\newcommandx{\change}[2][1=]{\todo[linecolor=red,backgroundcolor=red!25,bordercolor=red,#1]{Change: #2}}
\newcommandx{\info}[2][1=]{\todo[linecolor=olive,backgroundcolor=olive!25,bordercolor=olive,#1]{Info: #2}}
\newcommandx{\improvement}[2][1=]{\todo[linecolor=purple,backgroundcolor=purple!25,bordercolor=purple,#1]{Improve: #2}}
\newcommandx{\cut}[2][1=]{\todo[linecolor=yellow,backgroundcolor=yellow!25,bordercolor=yellow,#1]{Potential Cut: #2}}
\newcommand\todoin[2][]{\todo[inline, caption={Outline}, linecolor=green,backgroundcolor=green!25,bordercolor=green,size=small #1]{
\begin{minipage}{\textwidth-4pt}Outline: #2\end{minipage}}}

\newcommand{\itemtodo}[0]{\item[$\Box$]}

\newcommand{\attacks}[0]{\rightsquigarrow}
\newcommand{\supports}[0]{\rightarrow}
\newcommand{\args}[0]{\mathcal{A}}
\newcommand{\edges}[0]{\mathcal{E}}
\newcommand{\basescore}[0]{\tau}
\newcommand{\basescorex}[0]{\tau_{x}}
\newcommand{\edgeweights}[0]{w}
\newcommand{\targetargs}[0]{\mathcal{T}_{\delta}}
\newcommand{\Tnorm}[0]{T}
\newcommand{\Anorm}[0]{A_{\Tnorm}}

\newcommand{\edgeweightsstrict}[0]{w_{\succ}}

\newcommand{\edgeweightsirrelevant}[0]{w_{\nsim}}
\newcommand{\edgeweightsattacks}[0]{w_{\attacks}}
\newcommand{\edgeweightssupports}[0]{w_{\supports}}
\newcommand{\edgeweightsminimal}[0]{w_{c}}

\newcommand{\fullcasebase}[0]{\args'}

\newcommand{\casebasefilter}[2]{#1_{#2}}

\usepackage{bm}

\newcommand{\x}[1]{x_{#1}}
\newcommand{\y}[1]{y_{#1}}
\newcommand{\ypred}[0]{\hat y}

\newcommand{\argalpha}{i}
\newcommand{\argbeta}{j}
\newcommand{\arggamma}{k}

\newcommand{\xalpha}{\x{\argalpha}}
\newcommand{\xbeta}{\x{\argbeta}}
\newcommand{\xgamma}{\x{\arggamma}}

\newcommand{\yalpha}{\y{\argalpha}}
\newcommand{\ybeta}{\y{\argbeta}}

\newcommand{\batch}{\mathcal{B}_k}

\newcommand{\case}[2]{(\x{#1}, \y{#2})}
\newcommand{\casealpha}{\case{\argalpha}{\argalpha}}

\newcommand{\argnew}{N}
\newcommand{\casenew}{\case{\argnew}{?}}
\newcommand{\xnew}{\x{\argnew}}

\newcommand{\strength}[2]{\psi^{(#1)}_{#2}}
\newcommand{\aggregate}[2]{\rho^{(#1)}_{#2}}

\newcommand{\finalstrength}[0]{\sigma}

\newcommand{\loss}[0]{\mathcal{L}}

\newtheorem{definition}{Definition}[section]

\newcommand{\C}[1]{\mathcal{C}_{#1}}
\newcommand{\Calpha}[0]{\C{\argalpha}}
\newcommand{\Cbeta}[0]{\C{\argbeta}}
\newcommand{\Cgamma}[0]{\C{\arggamma}}
\newcommand{\Cnew}[0]{\C{\argnew}}
\newcommand{\Ctarget}[1]{\C{#1}^{\delta}}

\newcommand{\NN}[1]{f_{#1}}
\newcommand{\param}[1]{\theta_{#1}}

\newcommand{\parambs}[0]{\param{\tau}}
\newcommand{\paramw}[0]{\param{w}}

\newcommand{\NNbs}[0]{\NN{\parambs}}
\newcommand{\NNw}[0]{\NN{\paramw}}

\newcommand{\AM}[0]{A}
\newcommand{\A}[1]{\AM_{#1}}
\newcommand{\Acb}[0]{\A{cb}}
\newcommand{\AN}[0]{\A{new}}

\tikzset{
    attack/.style={line width=2.2pt, draw=red!90!black, -{Latex[length=2mm, width=2.5mm]}},
    support/.style={line width=2.2pt, draw=green!60!black, -{Latex[length=2mm, width=2.5mm]}}
}

%% file: introduction/introduction.tex


As deep learning becomes the dominant AI approach for high capacity, scaleable models across diverse data modalities, questions arise as to how deep learning models arrive at their predictions\cite{NN-interpretability,a-survey-on-nn-interpretability}. This becomes exceedingly difficult as deep neural networks (DNNs) scale, because the sheer volume of parameters and varying input-output behaviours render DNNs intractable to fully interpret~\cite{a-survey-on-nn-interpretability}. Additionally, the reasoning mechanism used by DNNs is not explicit. Often DNNs are trained with a task-specific objective end-to-end, meaning that their reasoning mechanism is tightly coupled with the latent features extracted. Therefore, one cannot easily explain how the training data give rise to a classification beyond the features extracted. Given the complexity of deep learning, even these feature-based explanations may not be faithful~\cite{m4-xai-faithfullness-benchmark,keep-the-faith-faithful-explanations-cnn-for-cbr}. But faithfully understanding the reasoning behind predictions is important for preventing model failures. Additionally, users may have an expectation and right to an explanation for decisions made by autonomous systems~\cite{gdpr-right-to-exp}.

Neurosymbolic AI integrates symbol manipulation with DNNs to gain the benefits of explicit reasoning and constraint enforcement that symbolic AI excels at~\cite{neuro-symbolic-ai-3rd-wave,data-and-knowledge-driven-ai-nesy-survey}. 
We
introduce \textit{Deep Arguing}, a neurosymbolic model that classifies data points using formal argumentative semantics as studied in computational argumentation, in the spirit of  \citep{gradual-argumentation-properties,mlp-semantics}. Computational argumentation is highly effective for modelling defeasibility in data, handling conflicting information, representing uncertainty, and explaination~\cite{argumentative-xai-survey}, all properties leveraged by Deep Arguing. 
Like some existing computational argumentation approaches~\cite{mlp-semantics}, Deep Arguing relies upon specific argumentative structures whereby arguments are equipped with an intrinsic  strength, interact by dialectical relationships of attack or support,   and these relationships are weighted according to their magnitude. In the case of Deep Arguing these argumentative structures are learnt, wherein data points are individual arguments. DNNs parametrise the intrinsic strength of 
arguments and the magnitude of the attack and support interactions. While learnable parameters reside within the DNNs, as their outputs induce the strength of arguments and the weights of their relationships, varying these parameters gives rise to different argumentative structures. Using differentiable argumentation semantics and careful consideration of soft constraints, the DNNs can be trained end-to-end through the argumentative structure. 

\begin{figure}
    \centering
    \includegraphics[width=1\linewidth]{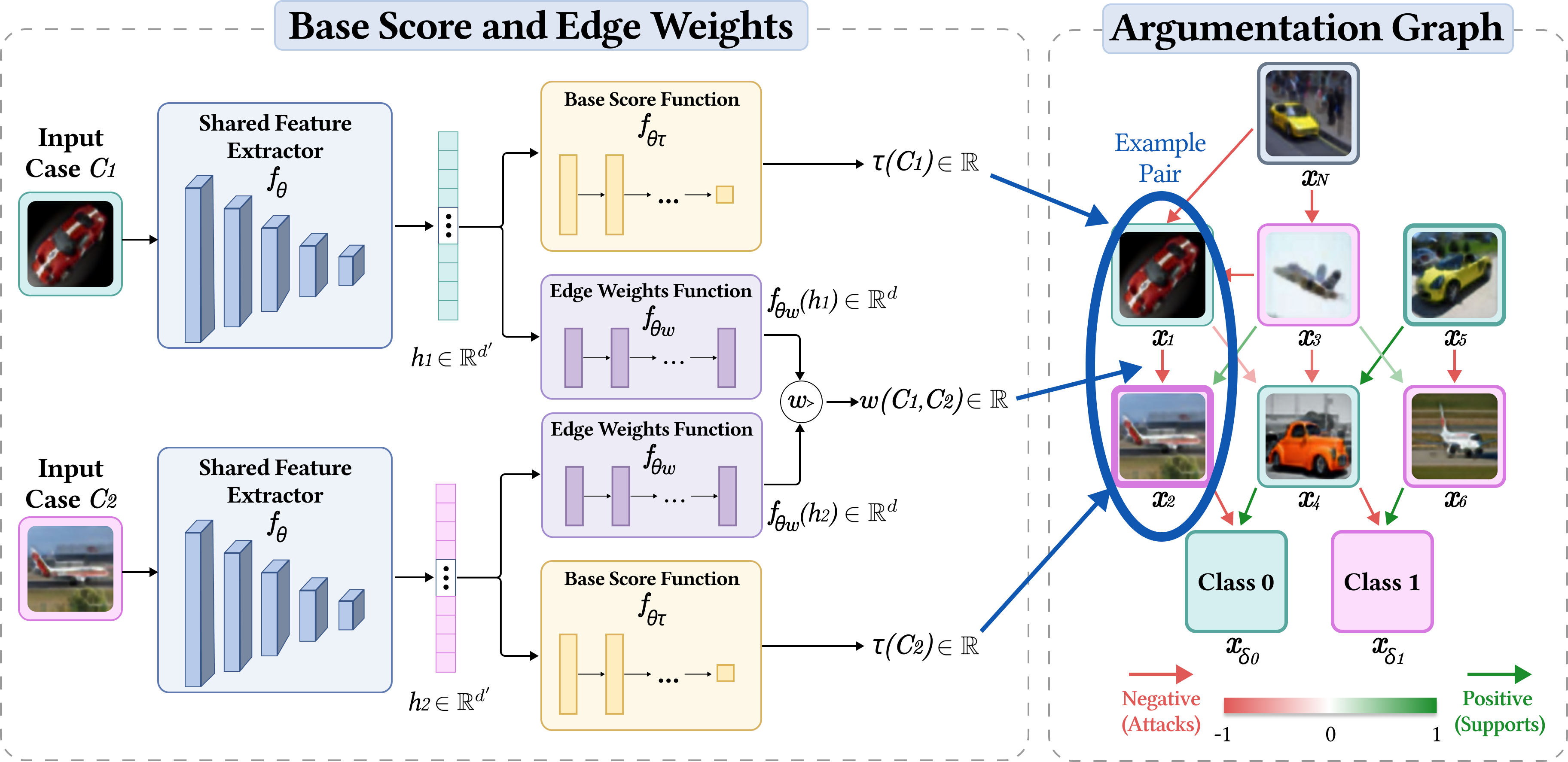}
    \caption{Left
    : 
    Illustrated architecture of parametrised base score and edge weight functions using a shared ResNet~\cite{resnet} feature extractor for two example cases of 
    CIFAR-10
    ~\cite{cifar-10}. Right
    : An example subset of a generated argumentation graph. Arguments are 
    images from 
    CIFAR-10, each coloured by its label
    , 
    with 
    the new case in grey. Arguments attack (or support) those with opposing (or equal) outcomes. For the example pair, we see the base scores (represented by border thickness) as well as the 
    weight and direction of the attack relation as computed by the parametrised functions.}
    \label{fig:architecture}
\end{figure}

As Deep Arguing does not directly modify the structure of the DNNs, one can select any architecture that best suits their use case, for example 
Multi-Layer Perceptrons (MLPs) for tabular data and convolutional neural networks (CNNs)~\cite{cnn-training} or 
ResNets~\cite{resnet} (as in Figure~\ref{fig:architecture}) for images. 
No matter which DNN architecture is used to underpin Deep Arguing, it results in an interpretable graph (see Figure~\ref{fig:cifar10-example} for an illustration) 
from which we can draw faithful case-based explanations that justify how the training data gives rise to a prediction.

In summary, our contributions are as follows.
%
(1)
We introduce Deep Arguing 
    as a novel neurosymbolic method for interpretable classification, integrating formal argumentative reasoning in conjunction with deep learning to enable structured and interpretable predictions.
 (2) We
 show that the framework 
    is constrainable through fuzzy logic and optimisation objectives, enabling desirable properties (e.g. minimality, acyclicity, sparsity
    ) that support both interpretability and scalable learning.
(3) We
conduct an empirical evaluation across tabular and imaging 
datasets, demonstrating that Deep Arguing achieves performance competitive with DNN baselines, while simultaneously providing interpretable argumentative reasoning.
(4) We 
carry out a case study on the CIFAR-10 dataset~\cite{cifar-10} providing insight into 
    Deep Arguing's reasoning process, illustrating how argumentation supports class discrimination and justifies  predictions between competing classes.

%% file: introduction/related.tex
\paragraph{Neurosymbolic AI} Neurosymbolic AI~\cite{neuro-symbolic-ai-3rd-wave,data-and-knowledge-driven-ai-nesy-survey} integrates deep learning with symbolic reasoning to achieve the flexibility and scalability of DNNs with interpretability. Often DNNs are used as perception modules (e.g. classifying MNIST digits) whilst the symbolic layer handles downstream reasoning (e.g. adding two digits)~\cite{deepproblog}. Many approaches encode reasoning using hand-crafted rules as part of the loss function~\cite{semantic-loss,ltn,deepproblog}. In these lines of work, although the reasoning layer may be interpretable, the underlying neural predictions remain opaque and weak supervision can lead to reasoning shortcuts~\cite{analysis-and-mitigation-reasoning-shortcuts}, where incorrect class assignments still lead to correct downstream outputs. Deep Arguing is closer to methods that jointly learn perception and reasoning~\cite{DSL,NSIL}, but these approaches may inherit the aforementioned issues. Instead Deep Arguing is fully supervised and addresses interpretable classification. Existing approaches often encode logic into model weights~\cite{RLL,MLLP,DDLGN}, which can be difficult to scale to complex problems given the rigidity of logical rules, restrictions to monotonic logics and applications to only MLP layers and not other architectures like CNNs. In contrast, Deep Arguing uses DNN to learn the dialectical relationships between data points as arguments, creating a graphical structure that provides a flexible inductive bias that supports non-monotonicity and defeasible reasoning~\cite{argumentative-xai-survey,DUNG-aa}.  

\paragraph{Neuro-Argumentative Learning} Previous attempts to combine DNNs with 
argumentation~\cite{neuro-argumentative-learning} often operate in a pipeline fashion in which the DNN is first trained for feature extraction and then the argumentation layer is used 
for the final classification~\cite{ANNA,DEAr,arg-llms,slot-attention-plus-aacbr}. Multi-layer perceptrons (MLPs) can also be interpreted as argumentation frameworks~\cite{mlp-semantics,sparx,af-to-nn1,af-to-nn2}, and have been used for interpretable prototypical learning~\cite{proto-arg-net}. However, this interpretation does not translate the representation of hidden neurons to meaningful concepts. Like Gradual AA-CBR~\cite{neuro-argumentative-learning}, our approach learns argumentation structures end-to-end with case-based reasoning for interpretability, but Deep Arguing improves gradient flow for faster convergence, better scaling, and reduced risk of suboptimal minima, thus allowing for the use of DNNs as the neural backbone for more complex datasets. It also stabilises argumentation semantics by preventing pairwise cycles between similar arguments and better optimisation of the graph structure, unlike Gradual AA-CBR. Furthermore, by reducing casebase size through clustering and avoiding large multiplicative aggregations, Deep Arguing achieves stronger gradients and fewer arguments, improving both learning efficiency and interpretability.

\paragraph{Graph Neural Networks} The formal 
semantics used by Deep Arguing resemble message passing in Graph Neural Networks (GNNs)~\cite{GNNs} but differ fundamentally in how information is represented and propagated. GNNs perform iterative updates in 
latent space using learnable transformations, producing representations that are not directly interpretable~\cite{gcns,gats}. Instead, Deep Arguing first compares latent feature representations to derive scalar values that represent semantically meaningful argument strengths and relationships. Update rules operate with these interpretable quantities rather than in a fully latent space.  Additionally, GNNs often aggregate information from a small number of message-passing layers~\cite{gcns,gats}, which can  limit their long range dependencies, whereas argumentation semantics can allow for interaction structures across the whole graph. Furthermore, GNN-based methods assume a predefined graph structure which may not be available for non-relational datasets~\cite{GNNs}. Deep Arguing instead learns argumentation graphs from provided data in a supervised manner. This is related to graph structure learning~\cite{gsl-survey} but differs in that the learned structure is explicitly tied to argumentation semantics rather than serving as input to a GNN.




%% file: preliminaries/preliminaries.tex

\paragraph{Edge-Weighted Quantitative Bipolar Argumentation Frameworks (QBAFs)}
A QBAF is a structure for representing arguments, their intrinsic strengths and the strength of the relationships between arguments ~\cite{mlp-semantics}. Formally,
a QBAF is a quadruple $\langle \args, \edges, \basescore, \edgeweights \rangle$, where $\args$ is a set, whose elements are 
\emph{arguments},\footnote{We assume that $\args = \{1, 2, \ldots, n\}$, that is the names of arguments are numbers for simpler presentation.} $\edges\ \subseteq \args \times \args$ 
are 
edges between arguments, $\tau: \args \rightarrow [0, 1]$ is a total function that maps every argument $\argalpha \in \args$ to a \emph{base score} and $\edgeweights: \edges \rightarrow \mathbb{R}$ is a total function that maps every \emph{edge} to a \emph{weight} 
(with a slight abuse, 
we write $\edgeweights(a, b)$ instead of $\edgeweights((a, b))$).
%
%
The {base scores} denote prior beliefs about the intrinsic strength of  arguments.  We deem negative {edge weights} 
to indicate \textit{attack}, 
positive ones to indicate \textit{support} and 
a zero value to be 
equivalent to no edge
. 
Each QBAF can be seen as a graph with arguments as nodes and supports and attacks as (differently coloured) edges (see Figure~\ref{fig:architecture}). 
MLPs can be interpreted as QBAFs~\cite{mlp-semantics}, whereby base scores correspond to biases. 

QBAFs can be used to compute the \textit{final strength} of each argument, representing the degree to which an argument is considered acceptable given the interactions between arguments in the graph and their base scores.
The final strengths can be computed using gradual semantics~\cite{gradual-semantics,gradual-argumentation-properties}, represented by the function $\finalstrength: \args \rightarrow [0,1]$. Multiple such gradual semantics exist, but we 
leverage the MLP-based Semantics~\cite{mlp-semantics}. Formally, 
    let $\strength{t}{\argalpha} \in [0,1]$ be the strength of argument $\argalpha$ at iteration~$t$. For every argument $\argalpha \in \args$, let $\strength{0}{\argalpha} := \basescore(\argalpha)$. The strength values are computed by iterative application of 
    \textit{aggregation}, defined as $\aggregate{t+1}{\argalpha} := \sum_{(\argbeta, \argalpha) \in \edges} \edgeweights(\argbeta, \argalpha) \cdot \strength{t}{b} $ and \textit{influence} as $\strength{t + 1}{\argalpha} := ReLU((\basescore(\argalpha) + \aggregate{t+1}{\argalpha})$.
The \textit{final strength} of argument $\argalpha$ is given by $\finalstrength(\argalpha) = \lim_{k \rightarrow \infty}\strength{k}{\argalpha}$ if the limit exists and $\bot$ otherwise. 
This semantics was originally defined to capture the ``reasoning'' of MLPs towards their predictions. Deep Arguing uses this semantics in a different context, reasoning over QBAFs that are not MLPs. 


\paragraph{Fuzzy Logic}
Fuzzy logics are real-valued approximations of logical languages in which propositions are in the range $[0, 1]$ where $0$ signifies false and $1$ true. \textit{T-norms} extend the notion of the logical conjunction $a \wedge b$. Formally, a T-norm is any function $\Tnorm: [0, 1]^2 \rightarrow [0, 1]$ that is commutative, associative, monotonic and satisfies $\Tnorm(1, a) = 1$~\cite{analysing-differentiable-fuzzy-logic-ops}. Common T-norms include the Gödel (minimum), product and Łukasiewicz T-norms~\cite{analysing-differentiable-fuzzy-logic-ops}. A \textit{fuzzy aggregation} operation extends the notion of the universal quantifier $\forall$. Formally fuzzy aggregation is a function $\Anorm: \bigcup_{n \in \mathbb{N}} [0, 1]^{n} \rightarrow [0, 1]$ that is constructed recursively by repeated application of a T-norm where $\Anorm() = 1$ and $\Anorm(x_1, \ldots, x_n) = \Tnorm(x_1, \Anorm(x_2, \ldots, x_n))$~\cite{analysing-differentiable-fuzzy-logic-ops}. Logical negation $¬a$ is expressed by strong negation as $1 - a$ and is used without explicit function representation throughout this work.\footnote{We do not specify fuzzy approximations of other common logical operations as they are not used in this work.
}

%% file: methodology/methodology.tex
\input{methodology/deep-arguing}

%% file: methodology/deep-arguing.tex
\textit{Deep Arguing} is a classification approach in the spirit of \cite{DEAr,neuro-argumentative-learning-with-cbr,aa-cbr,preference-based-aacbr}, whereby labelled samples (also called  \textit{cases}) interact through formal argumentation semantics about what label to assign to a \textit{new case}. The collection of labelled cases is called a \textit{casebase}. An argumentation structure is created from training data wherein cases argue in favour of their label and interact through \textit{attack} and \textit{support} relationships. 
Cases can only attack those with opposing labels and support those with agreeing labels (as illustrated in Figure~\ref{fig:architecture}, right, for the CIFAR-10 dataset~\cite{cifar-10}). 
The base score function $\basescore$ is used to weigh argument contributions. Similarly, the edge weights function $\edgeweights$ is used to determine how strongly arguments should attack or support others. The key idea is that cases that are considered more \textit{exceptional} attack or support those that are less 
so, thus creating a defeasible reasoning structure. 

Any unlabelled new case for which a prediction is sought, then attacks all cases which are \textit{irrelevant} to it, lessening their impact. For each class, there is one \textit{target argument}. Target arguments act as the topics of 
debate. After applying the argumentation semantics,  the class of the target argument with the greatest final strength 
is predicted for the new case. 

The base score and edge weight functions can be defined with learnable parameters. If the semantics used are differentiable, the parameters can be updated via gradient descent and backpropagation, meaning these functions can be represented by DNNs.
The specific architecture is task-dependant, for example we may choose a CNN for an image classification task or a MLP for tabular data. 
Figure~\ref{fig:architecture}, left, illustrates one such example architecture. Section~\ref{sec:experiments} discusses the specific choices for each dataset we experiment with. Formally, Deep Arguing is defined under the following domains: 


\begin{definition}[Deep Arguing Domain]
    \label{def:domain}
    Let $D_{cb} \subseteq X \times Y$ be a finite \emph{casebase} of labelled examples where $X$ is a set of \emph{characterisations} and $Y = \{1, 
    \ldots, C\}$ ($C \geq 2$) is 
    a set of possible \emph{outcomes}. Each case is of the form $\Calpha = \casealpha$
    . We define $\targetargs = \{(\x{c}^{\delta}, c)~|~c \in Y \}$ as the set of \emph{target arguments}, $\Ctarget{c}=(\x{c}^{\delta}, c)$, each one corresponding to exactly one class label, with $\x{c}^{\delta}$ the \emph{default characterisation} for class $c$. We let $\fullcasebase = D_{cb} \cup \targetargs$ be the \emph{full set of labelled cases} and $\casebasefilter{\fullcasebase}{c}  = \{ (x, y) \in \fullcasebase~|~c = y \}$ be the subset of $\fullcasebase$ containing only and all cases with label $c$.  
\end{definition}
Given any such domain, Deep Arguing Argumentation Graphs are as follows:
\begin{definition}[Deep Arguing Argumentation Graphs]
    \label{def:deep-arguing}
    Let $\basescorex:\! X \rightarrow \![0, 1]$ be a function mapping characterisations to base scores.
    Let $\edgeweightsstrict\!: \!X \times X \rightarrow [0, 1]$ and $\edgeweightsirrelevant: X \times X \rightarrow [-1, 0]$ represent the degree of \textit{exceptionality} and \textit{irrelevance} between input characterisations (respectively for cases and new cases). Let $\Tnorm: [0, 1]^{2} \rightarrow [0, 1]$ be a T-norm and $\Anorm: \bigcup_{n \in \mathbb{N}} [0, 1]^{n} \rightarrow [0, 1]$ be a fuzzy aggregation operator. Let $\Cnew = \casenew$ be a \emph{new case} with $x_n \in X$ a characterisation and $y_{?}\not\in Y$ an unknown outcome. 
    The \emph{QBAF mined from} $\Cnew$ and casebase $D_{cb}$ is $QBAF(D_{cb}, \Cnew) = \langle\args, \edges, \basescore, \edgeweights\rangle$ with:

    \begin{itemize}
        \item $\args = \fullcasebase \cup \{\Cnew\}$, \hfill (arguments)
        
        \item $\basescore(\Calpha) = \basescorex(\x{\argalpha})$, \hfill (base scores)
        
        \item $\edges = 
        \{ (\Calpha, \Cbeta) \in \fullcasebase\}~|~\argalpha \not = \argbeta \} \cup 
        \{ (\Cnew, \Calpha)~|~\Calpha \in \fullcasebase \}$, \hfill (edges) 

        \item $\edgeweights(\Calpha, \Cbeta)$ = 
        $
        \begin{cases}
            \edgeweightsirrelevant(\xnew, \xbeta) & \text{if } \Calpha = \Cnew,\\
            \edgeweightsattacks(\Calpha, \Cbeta)  & \text{if } \yalpha \not = \ybeta,\\
            \edgeweightssupports(\Calpha, \Cbeta) & \text{otherwise, \, where:}
        \end{cases}
        $\hfill (edge weights)

    \begin{itemize}[label=$\circ$]
    
        \item $\edgeweightsattacks(\Calpha, \Cbeta) = -\edgeweightsminimal(\Calpha, \Cbeta, \casebasefilter{\fullcasebase}{ \yalpha})$, \hfill (attacks)
        
        \item $\edgeweightssupports(\Calpha, \Cbeta) = \edgeweightsminimal(\Calpha, \Cbeta, \fullcasebase)$, \hfill (supports)
        \item $\edgeweightsminimal(\Calpha, \Cbeta, S) = \Tnorm\Big(\edgeweightsstrict(\xalpha, \xbeta), \Anorm(w_m(\Calpha, \Cbeta, \Cgamma)_{\Cgamma \in S}) \Big)$ 
        \hfill (exceptionality)
        \item $w_m(\Calpha, \Cbeta, \Cgamma) = 1 - \Tnorm\Big{(}\edgeweightsstrict(\xalpha, \xgamma), \edgeweightsstrict(\xgamma, \xbeta)\Big{)}$ 
        \hfill (minimality)
            
        \end{itemize}
    \end{itemize}
\end{definition}
Figure~\ref{fig:cifar10-example} gives an example of a subgraph of an induced QBAF. Whilst we define edges for all pairs of arguments in the casebase, the edge weight function can output edges with a value of $0$ which is semantically the same as no edge between the arguments, so we choose not to visualise these edges. We can use the induced QBAF to make predictions for a new case as follows:

\begin{definition}[Deep Arguing Prediction]
\label{def:prediction}
The label for new case $\Cnew$ is predicted as follows:
    $$\textnormal{DeepArguing}(D_{cb}, \Cnew) = \underset{c \in Y}{\textnormal{argmax}} \,  \finalstrength(\Ctarget{c}),$$
where $\finalstrength$ is a gradual semantics computing the final strength of the arguments in the QBAF mined from $D$ and $\Cnew$. 
\end{definition}
Intuitively, the class of the target argument with the maximum final strength is predicted for 
$\Cnew$.

To ensure coherent and interpretable reasoning we enforce additional constraints when learning edge weight functions. Specifically, in Section~\ref{sec:min}, a \textit{minimality} condition enforces that arguments of least intervening exceptionality have larger magnitude edge weights relative to those of greater exceptionality. This condition is defined with fuzzy logic, using soft differentiable logical operations~\cite{analysing-differentiable-fuzzy-logic-ops}. Additionally, in Section~\ref{sec:methodology:functions} we define edge-weight functions that prevent pairwise cycles. Moreover, in Section~\ref{sec:training:loss}, we use loss terms which can be used in training to guide the learning, for example to 
enforce that the learned mined argumentation graph is acyclic.

\subsection{Enforcing minimality with fuzzy logic constraints}
\label{sec:min}

By interpreting the T-norm operations as a logical AND operator and the 
fuzzy aggregator operator as the $\forall$-quantifier, the exceptionality and minimality constraints in Definition~\ref{def:deep-arguing} allow 
an edge from $\Calpha$ to $\Cbeta$ 
to be strongly weighted only when $\Calpha$ is more exceptional than $\Cbeta$ and there is no intermediate case $\Cgamma$ in some set $S$, that is more exceptional than $\Cbeta$ but less exceptional than $\Calpha$. In other words, edges are strongly weighted only between minimally separate cases in the exceptionality ordering relative to $S$. The scenarios in which minimality applies is described in the supplementary material.


To enforce the minimality condition whilst still allowing differentiable learning of $\basescorex$ and $\edgeweightsstrict$, we use the Gödel T-norm $\Tnorm(a,b)=min(a,b)$. Since the aggregation operator $\Anorm$ implements a universal quantification over a set of cases and therefore involves a minimum over multiple values, we replace it with a smooth relaxation based on LogSumExp~\cite{logsumexp}: $\Anorm(a_1, \ldots, a_n) = -t\log\sum_{i = 1}^{n}e^{-a_{i}/t}$. As $t \rightarrow 0$, this expression converges to $\min(a_1, \ldots a_n)$, while for finite $t$ it provides smoother gradients during optimisation. In practice, $t$ is treated as a small fixed temperature hyperparameter.


\subsection{Edge-weight and base score functions}
\label{sec:methodology:functions}

For the base score function, we let $\basescorex(\x{\argalpha})  = \NNbs(\xalpha)$ where $\NNbs$ is a DNN with learnable parameters $\parambs$. This means that a DNN is used to directly output the base score for each argument, determining their intrinsic importance in the QBAF.

For the edge weight function, we let \\
\hspace*{1cm} $\edgeweightsstrict(\xalpha, \xbeta) = \text{ReLU}\bigg{(}\frac{1}{d}\sum_{l=0}^{d}\bigg{[} \text{sigmoid}\Big{(}\alpha\cdot(\NNw(\xalpha)_{l} - \NNw(\xbeta)_{l})\Big{)} \cdot 2 - 1\bigg{]}\bigg{)}$
\\
where $\NNw$ is a DNN with learnable parameters $\paramw$, $\alpha$ is a temperature hyperparameter, and $\NNw(\xalpha) \in \mathbb{R}^{d}$ returns a feature vector of dimensionality $d$ which is a hyperparameter. First, for each input we compute a $d$-dimensional latent feature representation. Then for each feature, we compute the difference between them, rescaled by sigmoid to emit a value in the range $[0, 1]$. In this instance a value near $1$ means that the feature extracted from $\xalpha$ is larger than $\xbeta$, near $0$ means the reverse is true and a value near $0.5$ means the features are approximately the same. We want to average this value over all features giving a general score of exceptionality. However, if we just average over this value, we may get pair-wise edges $(\Calpha,\Cbeta)$ and $(\Cbeta,\Calpha)$ 
wherein $\Calpha$ attacks (or supports) $\Cbeta$ and $\Cbeta$ attacks (or supports) $\Calpha$ which can prevent convergence of the argumentation semantics~\cite{mlp-semantics}. Thus, we rescale the function and apply a ReLU so that it now outputs a value in the range $[0, 1]$ representing the degree to which $\xalpha$ (i.e., the characterisation of $\Calpha$) is more exceptional than $\xbeta$ (i.e., the characterisation of $\Cbeta$) or $0$ if $\xalpha$ is not more exceptional than $\xbeta$ (i.e. the feature values are on average the same). Thus the edge weight function acts as a one-way soft coordinate domination function in the latent feature space to disallow pairwise cycles. 



For the irrelevance function, we let $\edgeweightsirrelevant(\xnew, \xbeta) = 1 - \edgeweightsstrict(\xnew, \xbeta)$. This ensures that any cases 
more exceptional than the new case are irrelevant to the new case and thus attacked by it. Additionally, by defining irrelevance in terms of exceptionality, we ensure the two concepts do not learn separate representations that could otherwise result in learning shortcuts that do not leverage argumentative reasoning.

The architecture choice for each DNN is dependent on the data type, the number of input features and the required learning capacity for the specified task as discussed in Section~\ref{sec:experiments}. Moreover, $\parambs$ and $\paramw$ need not be mutually exclusive. For example, with the CIFAR-10 dataset~\cite{cifar-10}, $\NNbs(\xalpha) = f_{\parambs'}(ResNet_{\theta}(\xalpha))$ and $\NNw(\xalpha) = f_{\paramw'}(ResNet_{\theta}(\xalpha))$ where $ResNet_{\theta}$~\cite{resnet} is a shared feature extractor, and $f_{\parambs'}$ and $f_{\paramw'}$ are MLPs as shown in Figure~\ref{fig:architecture}.

\section{Training}
\label{sec:training}




In this section, we conclude the presentation of Deep Arguing by laying out the details of training,  algorithmically in Section \ref{sec:training:algorithm} and in terms of the optimisation objectives in Section \ref{sec:training:loss}. 

\subsection{Training algorithm}
\label{sec:training:algorithm}

Deep Arguing is trained using a gradient descent algorithm, 
given in the supplementary material. 
The key difference in training Deep Arguing is that it contains a fit step in the forward pass in which the QBAF is constructed as described by Definition~\ref{def:deep-arguing}. It is during this construction 
that there are calls to $\basescorex$ and $\edgeweightsstrict$, hence this fit step is defined using only differentiable operations. During the backward pass, the parameters of  $\basescorex$ and $\edgeweightsstrict$ are updated, which can be viewed as learning the reasoning structure required to solve the task. Arguments and relationships unimportant for the classification will have base scores and edge weights near $0$ whilst more salient ones will be near $1$.
Some or all of the parameters of $\basescorex$  and $\edgeweightsstrict$ may be pre-trained. For example, if using a shared feature extractor, we may first wish to pre-train this 
to allow the model to learn initial feature representations that can then be updated later when combined with the base score and edge weight functions. 

We can define the casebase using all the data points in the training dataset, 
but this 
may lead to large QBAFs that are computationally expensive to generate, lead to slow learning convergence and are cognitively intractable for humans to parse~\cite{neuro-argumentative-learning-with-cbr}. Instead, the casebase can be first constructed using a clustering technique. We use K-Means clustering~\cite{k-means-1,k-means-2}, taking the data points nearest the cluster centres as the casebase. The number of clusters per class is a hyperparameter. 

During training, for each new case the constructed QBAF is identical except for the associated argument and irrelevance edges
. Thus, in practice, we execute the fit stage once per batch on the casebase without considering the new cases and represent the resulting graph as an  adjacency matrix $\Acb \in \mathbb{R}^{n \times n}$ where $n$ is the size of the casebase. We then compute the base scores of the casebase, $\mathbf{b}_{cb} \in \mathbb{R}^{n}$, using vectorised operations. To compute the semantics for a batch of $B$ new cases efficiently, we tile the base scores across the batch dimension so that the strengths for all casebase arguments can be updated simultaneously. This tiling gives an initial strengths matrix $S^{(0)} \in \mathbb{R}^{B \times n}$. Similarly, we can represent the base scores of the new cases as $\mathbf{b}_{new} \in \mathbb{R}^{B}$ and the adjacency matrix of the attacks from the new cases as $\AN \in \mathbb{R}^{B \times n}$. We can then compute the aggregation step using $\AN$ with $\mathbf{b}_{new}$ as the initial strengths of the new cases, giving 
$S^{(1)}$ representing the strengths of casebase arguments after the effect from the new cases is considered. Aggregation is implemented as 
matrix multiplication. This formulation allows the semantics computation to be implemented entirely using differentiable linear algebra operations, enabling efficient gradient-based learning. As the new cases introduce only outgoing attack relations, they do not participate in cycles and so their effect can by applied in one step. We then iteratively apply the aggregation step using $\Acb$, starting with $S^{(1)}$ and the influence step using $\mathbf{b}_{cb}$ giving $S^{(i + 1)}$ from $S^{(i)}$ for a fixed number of steps $I$ which is a hyperparameter. We take $\sigma = S^{(I)}$ as the final strengths of all arguments across the entire batch.

We iterate for $I$ steps as convergence of the semantics is not guaranteed if the learned argumentation structure contains a cycle. Cyclic attack (or support) paths in a QBAF can cause an argument to indirectly undermine (or reinforce) itself, leading to oscillations in the semantics~\cite{Potyka18-convergence,extended-modular-semantics,mlp-semantics}. By constraining to $I$ steps, we stop iterating regardless of whether convergence occurs. As described in Section~\ref{sec:training:loss}, we optimise towards an acyclic graph, but cycles may occur as this is a soft constraint so we cannot assume acyclicity. Additionally, repeated multiplications between edge weights and strengths in $[0, 1]$ lead to vanishing gradients as the number of iterations increases. Setting a fixed number of iterations constrains the graph structure to propagate strengths from the new case to the target arguments (see Definition~\ref{def:domain}) with at most $I$ reasoning steps.

\subsection{Optimisation objective and soft constraints on the graph structure}
\label{sec:training:loss}


The loss function $\loss$ minimises a classification objective $\loss_{task}$ as well as additional loss terms that enforce soft constraints on the learned graph structure and edge weight functions. We let $\loss_{task}$ be a cross entropy loss function.\footnote{This 
is also the case for binary classification tasks as if we instead chose to have a single target argument to use a binary cross entropy loss then 
the choice inherently biases the model towards one class over the other~\cite{DEAr}.}As 
$\loss_{task}$
depends on the predictions made by the argumentative reasoning, one could interpret this as a semantic loss function akin to~\cite{semantic-loss,ltn,deepproblog} but using argumentation semantics and a learned argumentation structure rather than logic/logic programming semantics.

To ensure that $\edgeweightsstrict$ effectively models a notion of exceptionality, we use the loss term 
\[
\loss_{\delta}(\mathbf{x_{cb}}, \mathbf{x_{\delta}}) = \text{MSE}(\edgeweightsstrict(\mathbf{x_{cb}}, \mathbf{x_{\delta}}), 1) + \text{MSE}(\edgeweightsstrict(\mathbf{x_{\delta}}, \mathbf{x_{cb}}), 0)
\]
where $\text{MSE}$ is the mean-square error, $\mathbf{x_{cb}}$ represents all 
characterisation values of $\casealpha \in D_{cb}$ and analogous for $\mathbf{x_{\delta}}$ in $\targetargs$ (the set of target arguments, see Definition~\ref{def:domain}). This ensures that every case in the casebase is always considered more exceptional than the target arguments and 
these are never considered more exceptional than any casebase argument. Additionally, we set the characterisation of each target argument to the mean average of the casebase characterisation $\mathbf{x_{cb}}$. This exceptionality represents a learned soft partial order in the latent space anchored at the empirical mean of the input distribution; in other words the ordering reflects representation-level deviation from the dataset centre. 

Inspired by graph structure learning~\cite{gsl-survey}, further loss terms can be used to regularise the structure of the argumentation graph. As argumentation semantics is not guaranteed to converge for cyclic graphs~\cite{Potyka18-convergence,extended-modular-semantics,mlp-semantics}, we use a loss term to learn directed acyclic graphs (DAGs)~\cite{dag-notears}, let  
\[
\loss_{dag}(\AM) = tr(e^{\AM \circ \AM}) - n
\] 
where $\circ$ is the Hadamard product, $n$ is the size of the casebase and $e^{\AM}$ is the matrix exponential of 
$\AM$. By Theorem 1 of~\cite{dag-notears}, $\AM$ is guaranteed to be a DAG when this value is $0$. We only apply this to $\Acb$ as $\AN$ cannot contain cycles as it 
holds only attacks from the new case. To enforce sparsity we use 
\[
\loss_{sp}(\AM) = \frac{1}{n} \sum_{i,j}|\AM_{i,j}|
\] which minimises an $L_{1}$ term scaled for the size ($n$) of the casebase. We apply this to both $\Acb$ and $\AN$ during training, which ensures that less important edges have the least weight. 

The full loss function is therefore:
\[
\loss = \loss_{task}(\mathbf{\ypred}, \mathbf{y}) + \lambda_{\delta}\loss_{\delta}(\mathbf{x_{cb}}, \mathbf{x_{\delta}})  + \lambda_{dag}\loss_{dag}(\Acb) + \lambda_{sp}\loss_{sp}(\Acb) + \lambda_{sp'}\loss_{sp}(\AN)
\]
where $\mathbf{\ypred}$ and $\mathbf{y}$ are the predictions and true labels of a batch of cases, and the coefficients $\lambda_{\delta}$, $\lambda_{dag}$, $\lambda_{sp}$ and $\lambda_{sp'}$ are real value hyperparameters controlling the weight of the constraint loss terms.

%% file: experiments/experiments.tex
\begin{table}
  \caption{Comparison of
  F1 scores (\% mean $\pm$ 1 std over 5 runs). Deep Arguing is evaluated against Gradual AA-CBR (gAA-CBR), a DNN backbone, and a Decision Tree (DT).
  \texttimes\ stands for timeout.}
  \label{tab:tabular-results}
  \centering
  \begin{tabular}{lllll}
    \toprule
    Dataset     & Deep Arguing  & gAA-CBR   & DNN & DT  \\
    \midrule
    Adult~\cite{adult} 
      & $78.78 \pm 0.75$  
      & \texttimes  
      & $79.01 \pm 0.45$ 
      & $78.06 \pm 0.02$    \\

    Bank~\cite{bank-marketing} 
      & $76.81 \pm 0.40$     
      & \texttimes  
      & $77.36 \pm 0.44$ 
      & $72.27 \pm 0.07$     \\

    Chess~\cite{chess} 
      & $75.93 \pm 2.06$     
      & \texttimes 
      & $82.42 \pm 1.33$ 
      & $56.54 \pm 0.47$  \\

    Covertype~\cite{covertype} 
      & $67.05 \pm 0.24$    
      & \texttimes  
      & $67.28 \pm 0.55$  
      & $54.16 \pm 190.16$     \\

    Glioma~\cite{glioma} 
      & $82.33 \pm 1.91$   
      & $34.10 \pm 1.06$  
      & $81.39 \pm 1.76$  
      & $82.02 \pm 0.45$     \\

    Higgs~\cite{higgs} 
      & $76.65 \pm 0.17$ 
      & \texttimes 
      & $77.89 \pm 0.03$ 
      & $70.37 \pm 0.00$      \\

    \midrule

    CIFAR-10~\cite{cifar-10} 
      & $83.95 \pm 0.21$ 
      & - 
      & $82.69 \pm 0.41$ 
      & -   \\

    MNIST~\cite{MNIST} 
      & $96.30 \pm 1.22$   
      & - 
      & $93.52 \pm 4.80$   
      & - \\

    Fash MNIST~\cite{fashionmnist} 
      & $84.83 \pm 1.04$  
      & - 
      & $81.41 \pm 5.35$ 
      & -   \\

    \bottomrule
  \end{tabular}
\end{table}


We experiment with classification tasks using Deep Arguing across two modalities. Specifically, we look at six challenging tabular data task and three imaging tasks. The datasets were selected to showcase how Deep Arguing is capable of scaling with limited performance trade-off compared to the underlying DNNs it uses. Details of the datasets 
are in the supplementary material. Moreover, we compare against Gradual AA-CBR~\cite{neuro-argumentative-learning-with-cbr}, as this is a similarly interpretable argumentation-based model. For the tabular data, we also compare against decision trees, 
as a purely symbolic 
model.  

For datasets where a separate test set was provided, we evaluated on that only and split the training data into  
$80\%/20\%$ for training/validation. For data where no test set was provided, $20\%$ of the data was used as the test set. The remaining data was then split again $80\%/20\%$ for training/validation. 

For each Deep Arguing instance, we employ a similar architecture in which there is first a DNN feature extractor followed by two smaller MLPs with an equal number of parameters for the base score and the edge weight models. For all tabular datasets, the feature extractor is an MLP, for MNIST and Fashion MNIST the feature extractor is a CNN and for CIFAR-10 the feature extractor is a ResNet~\cite{resnet}. All models are trained end-to-end with the argumentation module with the exception of the CIFAR-10 ResNet, in which we first pre-trained and then froze the ResNet and then trained just the base score and edge weight MLPs end-to-end, as this approach leads to faster training time. For the easier-to-classify datasets, hyperparameter tuning was done by hand. For more challenging datasets, hyperparameter tuning was completed with Tree-structured Parzen Estimator~\cite{TPE} 
implemented in Optuna~\cite{optuna}. Architecture and hyperparameter configurations are provided in the supplementary material. All models are implemented in PyTorch~\cite{pytorch} and trained with an AdamW optimiser~\cite{adamw}.  

For the Gradual AA-CBR baseline, applicable to tabular data only, we used the 
method 
in~\cite{neuro-argumentative-learning-with-cbr} 
using all cases in the training data for the casebase and with an MLP as the feature extractor that outputs a single value used for the edge weight computation and rescaled for the base score. The implementation was done in PyTorch using the same training set-up as 
for Deep Arguing. 

For the baseline 
DNNs, we 
used the same feature extractors as 
in the Deep Arguing models to ensure a fair comparison. These feature extractors were trained using the same hyperparameter configurations as Deep Arguing, where applicable. A final linear classification layer was appended to each model to produce class predictions. All baseline DNNs were implemented in PyTorch and trained using the same optimisation set-up as 
for Deep Arguing, including the use of the AdamW optimizer.

For the decision tree baseline, applicable to tabular data only, we used the default implementation 
by scikit-learn~\cite{scikit-learn}, with 
maximum tree depth constrained to 10, 
to ensure a comparable level of model complexity: 
this permits up to $2^{10}$ nodes, 
already 
exceeding the number of nodes in Deep Arguing explanations ($k \cdot C$, where $k$ 
is the number of clusters per class and $C$ the number of classes). 


\paragraph{Results}
We report test-set results averaged over five seeds in Table~\ref{tab:tabular-results}, using macro F1-score ($\pm$ 1 standard deviation) selected due to class imbalance. Gradual AA-CBR failed on 5 of the 6 tabular datasets; on the remaining dataset, Glioma, Deep Arguing achieved substantially better performance. Further discussion as to why this could be is provided in the supplementary material.

Deep Arguing and DNNs considerably outperform decision tree models for tabular datasets.
Deep Arguing performs comparatively with the DNN baselines on the majority of the datasets, 
with generally no clear difference across data modalities (the only dataset in which Deep Arguing is not within $0.015$ of the DNN is for Chess, which is a particular difficult dataset, even for the decision tree baseline that Deep Arguing outperforms). 
However, compared to DNNs, we gain the interpretability benefits in which the reasoning process for the classification is revealed by the argumentation 
graph as described in Section~\ref{sec:interpretability}.

%% file: interpretability/interpretability.tex






\begin{figure}
    \centering
    \includegraphics[width=0.85\linewidth]{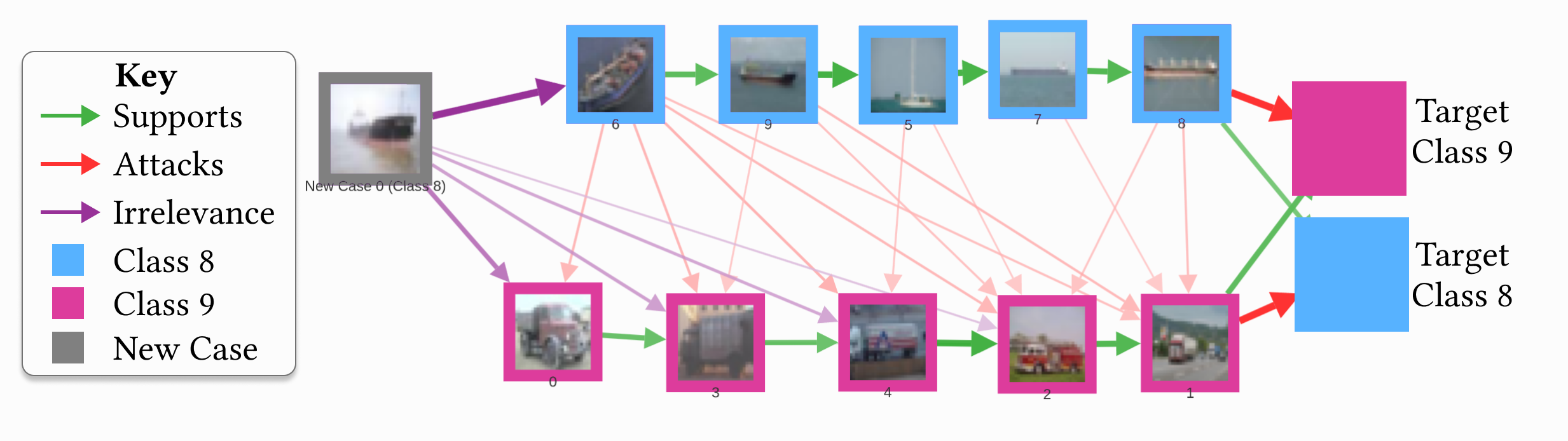}
    \caption{The subgraph of a QBAF for the CIFAR-10 dataset~\cite{cifar-10} filtered to classes 8 (ships) and 9 (trucks) and showing only edges with edge weights with a magnitude greater than 0.25. Edge weights are coloured by relationship type (irrelevance attacks behave as regular attacks, but are visualised in purple for clarity) with colour intensity representing the magnitude of the weights. Base scores are represented by border thickness. In this case all nodes have a base score of 1.0.}
    \label{fig:cifar10-example}
\end{figure}



Figure~\ref{fig:cifar10-example} illustrates a possible explanation drawn from a QBAF computed by Deep Arguing to classify 
a single new case of the CIFAR-10 dataset~\cite{cifar-10}.
This explanations is 
a subgraph constructed by filtering the full QBAF to classes of interest and edge weights above a set threshold. 
The full QBAF is provided in the supplementary material. 
%
The 
explanation shows that all but one of the trucks are considered irrelevant to the new case, whilst the only one ship that is photographed from an above angle is irrelevant to the new case. There are no attacks from trucks to ships, so the model has learned that the representation of ships are more exceptional than the representation of trucks. Because of these attacks, when the semantics is computed, the target argument for class 8 will have a final strength greater than the target argument for class 9. Thus the new case is classed as a ship. The 
explanation therefore provides a case-based reason for why a class is predicted and why not a different class. So although parameters of the edge weight and base score functions are learned from the entire training data, the reasoning process for the classification is defined over the casebase. 

The minimality condition ensures that not all arguments of the same label attack all those of opposing labels. For example we observe strong attacks towards the target arguments from two arguments of each class: these cases are the least exceptional for the class and thus the most representational for the images of that class. As these attack the target arguments, strong attacks from more exceptional cases to the target arguments would be redundant. Thus, minimality aids interpretability of the QBAF.

%% file: conclusion/conclusion.tex





We introduced Deep Arguing, a novel neurosymbolic model for interpretable classification, leveraging argumentative reasoning. We  showcased how to scale Deep Arguing
, by focusing on key optimisations of gradient flow and the argumentative structure. We showed empirically that Deep Arguing performs competitively with comparable DNN baselines and decision trees, where applicable.
We also provided a case study into the explanations Deep Arguing is capable of generating. 

{\bf Limitations}.
We focused on tabular data and images only, and plan to consider other data modalities and multi-modal settings as well as
data from high-stake domains such as healthcare, where 
it could be used to identify biases in models or data. 
Deep Arguing cannot leverage on background knowledge.
We plan to combine it with other neurosymbolic methods such as~\cite{semantic-loss,ltn,deepproblog} to 
complete tasks beyond classification that Deep Arguing is currently incapable of. Finally, human-specified preferences could be integrated into Deep Arguing akin to~\cite{preference-based-aacbr}, allowing for improved human-AI interactions.

%% file: appendix/minimality.tex
The minimality condition is described in Section~\ref{sec:min}. Here we provide an illustration of the scenarios where minimality applies.

For the attacks relation, the set $S$ contains only those cases with the same label as the potential attacker. To see why, consider three cases in decreasing order of exceptionality, $\Calpha, \Cbeta$ and $\Cgamma$. Suppose $\Calpha$ and $\Cbeta$ are labelled with class $1$ and $\Cgamma$ is labelled with class $2$. As depicted in Figure~\ref{fig:minimality-example-1}, in this setting it is redundant for both $\Calpha$ and $\Cbeta$ to attack $\Cgamma$; only the closest intervening exception should attack. Minimality therefore ensures that $\Cbeta$ will attack $\Cgamma$ with greater strength than $\Calpha$'s attack. If instead $\Cbeta$ had label $2$ as in Figure~\ref{fig:minimality-example-2}, then $\Calpha$ should attack both $\Cbeta$ and $\Cgamma$, since it represents an exception to each of them, which must be modelled. 

On the other hand, for the supports case, $S$ contains all cases regardless of the label. Suppose $\Calpha$, $\Cbeta$ and $\Cgamma$ each had the same label. Then a support from $\Calpha$ to $\Cgamma$ is redundant as $\Cgamma$ is already supported as in Figure~\ref{fig:minimality-example-3}. Now, suppose $\Calpha$ and $\Cgamma$ are in class $1$, whilst $\Cbeta$ is in class $2$ as in Figure~\ref{fig:minimality-example-4}. We would thus have $\Calpha$ attacks $\Cbeta$ which in turn attacks $\Cgamma$. So $\Calpha$ already indirectly defends $\Cgamma$ against its attack. An additional support relation between $\Calpha$ and $\Cgamma$ is redundant, so minimality prevents such edges from strong weighting. In effect, this definition of minimality allows exceptions represented by attacks to take priority over supports.

\newcommand{\figw}[0]{0.45}
\newcommand{\resize}[0]{0.6}

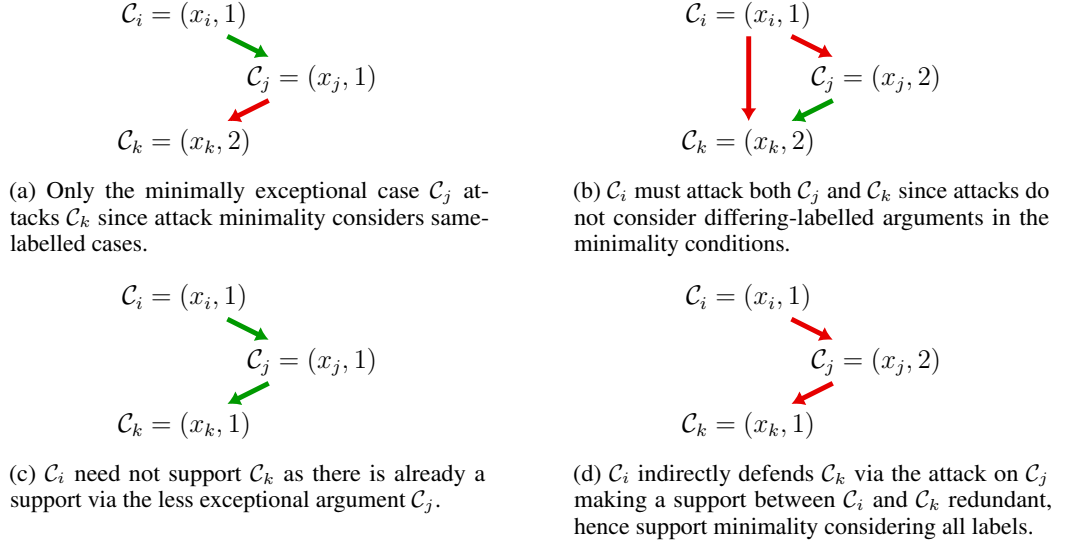
\begin{figure}
    \begin{subfigure}[t]{\figw\textwidth}
        \centering
        \resizebox{\resize\textwidth}{!}{
        \begin{tikzpicture}[main/.style = {draw=none, font=\large}]
    
            \node[main] (C0) at (-1, 2)  {$\Calpha = (\xalpha, 1)$};
            \node[main] (C1) at (1, 1)  {$\Cbeta = (\xbeta, 1)$};
            \node[main] (C2) at (-1, 0)  {$\Cgamma = (\xgamma, 2)$};

            \draw[attack] (C1) -- (C2);                   
            \draw[support] (C0) -- (C1);                   
            
            \begin{scope}[transparency group, opacity=0.5]
            \end{scope}
    
        \end{tikzpicture} 
        }
        \caption{Only the minimally exceptional case $\Cbeta$ attacks $\Cgamma$ since attack minimality considers same-labelled~cases.}
    \label{fig:minimality-example-1}
    \end{subfigure}
    \hspace{1cm}
    \begin{subfigure}[t]{\figw\textwidth}
        \centering
        \resizebox{\resize\textwidth}{!}{
        \begin{tikzpicture}[main/.style = {draw=none, font=\large}]
    
            \node[main] (C0) at (-1, 2)  {$\Calpha = (\xalpha, 1)$};
            \node[main] (C1) at (1, 1)  {$\Cbeta = (\xbeta, 2)$};
            \node[main] (C2) at (-1, 0)  {$\Cgamma = (\xgamma, 2)$};

            \draw[attack] (C0) -- (C1);                   
            \draw[attack] (C0) -- (C2);                   

            \draw[support] (C1) -- (C2);                   
            
            \begin{scope}[transparency group, opacity=0.5]
            \end{scope}
    
        \end{tikzpicture} 
        }
        \caption{$\Calpha$ must attack both $\Cbeta$ and $\Cgamma$ since attacks do not consider differing-labelled arguments in the minimality conditions.}
    \label{fig:minimality-example-2}
    \end{subfigure}
    \vspace{0.2cm}

    \begin{subfigure}[t]{\figw\textwidth}
        \centering
        \resizebox{\resize\textwidth}{!}{
        \begin{tikzpicture}[main/.style = {draw=none, font=\large}]
    
            \node[main] (C0) at (-1, 2)  {$\Calpha = (\xalpha, 1)$};
            \node[main] (C1) at (1, 1)  {$\Cbeta = (\xbeta, 1)$};
            \node[main] (C2) at (-1, 0)  {$\Cgamma = (\xgamma, 1)$};

            \draw[support] (C1) -- (C2);                   
            \draw[support] (C0) -- (C1);                   
            
            \begin{scope}[transparency group, opacity=0.5]
            \end{scope}
    
        \end{tikzpicture} 
        }
        \caption{$\Calpha$ need not support $\Cgamma$ as there is already a support via the less exceptional argument $\Cbeta$.}
    \label{fig:minimality-example-3}
    \end{subfigure}
    \hspace{1cm}
    \begin{subfigure}[t]{\figw\textwidth}
        \centering
        \resizebox{\resize\textwidth}{!}{
        \begin{tikzpicture}[main/.style = {draw=none, font=\large}]
    
            \node[main] (C0) at (-1, 2)  {$\Calpha = (\xalpha, 1)$};
            \node[main] (C1) at (1, 1)  {$\Cbeta = (\xbeta, 2)$};
            \node[main] (C2) at (-1, 0)  {$\Cgamma = (\xgamma, 1)$};

            \draw[attack] (C0) -- (C1);                   
            \draw[attack] (C1) -- (C2);                   

            \begin{scope}[transparency group, opacity=0.5]
            \end{scope}
    
        \end{tikzpicture} 
        }
        \caption{$\Calpha$ indirectly defends $\Cgamma$ via the attack on $\Cbeta$ making a support between $\Calpha$ and $\Cgamma$ redundant, hence support minimality considering all labels.}
    \label{fig:minimality-example-4}
    \end{subfigure}
    \caption{Examples illustrating the minimality constraint for attacks (in \textcolor{red!90!black}{red}) and supports (in \textcolor{green!60!black}{green}) given that $\xalpha$ is more exceptional than $\xbeta$ which is more exceptional than $\xgamma$. In practice this is a soft constraint so minimality will not entirely remove edges but rather lower their weight.}
    \label{fig:minimality-examples}
\end{figure}

%% file: appendix/training-algorithm.tex
\begin{algorithm}[t!]
\caption{Deep Arguing Training with Gradient Descent}
\label{alg:model-training}
\begin{algorithmic}[1]  
    \State \textbf{Input:} Training data $D_t$, learning rate $\lambda$, epochs $E$, batch size $B$, semantics $\sigma$, base score function $\basescorex$, edge weight functions $\edgeweightsstrict$ and $\edgeweightsirrelevant$, target arguments $\targetargs$
    \State \textbf{Initialize:} Parameters $\theta$ of $\basescorex$, $\edgeweightsstrict$, $\edgeweightsirrelevant$
    \State \textbf{Initialize:} Casebase $D_{cb}$ via clustering 
    \For{$\text{epoch} = 1$ to $E$}
        \State Shuffle $D_t$ and split into mini-batches $\mathcal{B}_1, \ldots, \mathcal{B}_K$ of size $B$
        \For{each mini-batch $\batch$}
            \ForAll{$\Cnew \in \batch$} \Comment{Operations are vectorised in practice}
                \State \textbf{Forward Pass:}
                \State Fit QBAF: $QBAF(D_{cb}, \Cnew)$ \label{alg:model-training:fit-step}
                \State Compute prediction $\ypred := \textnormal{DeepArguing}(D, \Cnew)$ \label{alg:model-training:semantics}
                \State Compute loss $\loss(\ypred, y)$
                \State \textbf{Backward Pass:}
                \State Compute gradient $\nabla_\theta \loss(\ypred, y)$
                \State Update: $\theta := \theta - \lambda \nabla_\theta \loss(\ypred, y)$
            \EndFor
        \EndFor
    \EndFor
    \State \textbf{Output:} Trained weights $\theta$
\end{algorithmic}
\end{algorithm}

Here we present the training algorithm for Deep Arguing. Algorithm~\ref{alg:model-training} demonstrates a gradient descent algorithm, however there is a key difference during the forward pass on line~\ref{alg:model-training:fit-step} in which there is a fit step that first constructs the QBAF with respect to $D_{cb}$ and $C_{N}$. Construction occurs according to Definition~\ref{def:deep-arguing}, wherein all operations are differentiable. On line~\ref{alg:model-training:semantics}, the predicted class is computed. This uses Definition~\ref{def:prediction} to compute the semantics of the QBAF. Again, the computation of the semantics is differentiable. This allows us to update the model weights using backpropagation unmodified. We make use of the PyTorch~\cite{pytorch}  for this implementation. More details on the training implementation are discussed in the main paper.

%% file: appendix/experiment-details.tex
\label{appendix:experiment-details}

\subsection{Datasets}
\label{appendix:datasets}

\begin{table}
  \caption{Properties of the datasets. CIFAR-10 and MNIST do not provide formal licenses but they are publicaly available and common for research/educational use.
  }
  \label{tab:data}
  \centering
  \begin{tabular}{lllll}
    \toprule
    Dataset     & no. instances  & no. classes   & no. features & Licence  \\
    \midrule
    Adult & 48842  & 2  & 14 & CC BY 4.0    \\
    Bank Marketing & 45211  & 2 & 16 & CC BY 4.0 \\
    Chess & 28056 & 18  & 6 & CC BY 4.0   \\
    Covertype & 581012 & 7   & 54 & CC BY 4.0 \\
    Glioma & 839 &  2 & 23 &  CC BY 4.0  \\
    Higgs &  11000000 & 2 &  27 & CC BY 4.0   \\
    \midrule
    CIFAR-10 & 60000 & 10 & 32 x 32 & -  \\
    MNIST &60000& 10 &28 x 28 & -    \\
    Fashion MNIST & 60000 & 10 & 28 x 28 & MIT  \\
    \bottomrule
  \end{tabular}
\end{table}

Table~\ref{tab:data} showcases the number of instances, number of classes and the number of features or dimensionality of the data, in the each dataset. We selected each tabular dataset because they contain multiple orders of magnitude more instances, and a far greater number of features than was evaluated in previous work on Gradual AA-CBR~\cite{neuro-argumentative-learning-with-cbr}. The choice of datasets demonstrates how Deep Arguing is capable of far greater scaling. 

All tabular datasets are available on the UCI ML repository: \url{https://archive.ics.uci.edu/}. CIFAR-10 is provided by the university of Toronto \url{https://www.cs.toronto.edu/~kriz/cifar.html}. MNIST and Fashion MNIST were provided by the Torchvisions library~\cite{pytorch}.

\subsection{Deep Arguing Model Architectures}

For all datasets, a DNN backbone was used as a feature extractor followed by an MLP for the base score and an MLP for the edge weight functions. The output of the feature extractor is passed into both MLPs. All MLPs use ReLU activation after each layer except the final output layers.  
Note that the DNN used in the baseline has the same architecture as the DNN feature extractor for Deep Arguing with a linear layer output for classification so we do not explicitly provide those architectures. 

\paragraph{Tabular Dataset}

All tabular datasets used a similar architecture with a different number of hidden neurons per layer. The feature extractor DNN was a two layer MLP, with the input layer for each feature extractor relative to the number of features (categorical features,  were one-hot encoded, so the size of the input layer may be greater than the number of features listed in Table~\ref{tab:data}) The number of neurons for each layer is shown in Table~\ref{tab:tabular-arch:fe}. The output of this MLP was then passed into an  MLP for the base score function and an MLP for the edge weights function. These use the same architecture except with different output layers. These were two (three for chess) layer MLPs with the number of neurons shown in Table~\ref{tab:tabular-arch:bs-ew}. The output for the base score function is a scalar value passed through a sigmoid activation function. The MLP edge weight output layer gives the $d$-dimensional feature vector as described in Section~\ref{sec:methodology:functions}.

\begin{table}
  \caption{ Number of neurons in the MLP feature extrator for the tabular datasets.
  }
  \label{tab:tabular-arch:fe}
  \centering
  \begin{tabular}{lll}
    \toprule
    Dataset     & MLP FE Hidden Layer  & MLP FE output layer    \\
    \midrule
    Adult & 64  & 64  \\
    Bank Marketing & 64  & 64  \\
    Chess & 128 & 64   \\
    Covertype & 512 & 256 \\
    Glioma & 64 &  64 \\
    Higgs &  512 & 256   \\
    \bottomrule
  \end{tabular}
\end{table}

\begin{table}
  \caption{ Number of neurons in the MLP for the bases score and edge weight functions
  }
  \label{tab:tabular-arch:bs-ew}
  \centering
  \begin{tabular}{llll}
    \toprule
    Dataset     & MLP Hidden Layer 1 & MLP Hidden Layer 2   & MLP edge weight output layer    \\
    \midrule
    Adult & 64  & - & 128 \\
    Bank Marketing & 64  & - & 64  \\
    Chess & 128 & 64 &  128 \\
    Covertype & 128 & - & 128 \\
    Glioma & 64 &  - & 64 \\
    Higgs & 128  & - & 128  \\
    \bottomrule
  \end{tabular}
\end{table}

\paragraph{MNIST and Fashion MNIST}

For both MNIST and Fashion-MNIST, the same architecture was used. The DNN feature extractor was a CNN consisting of three convolutional layers with kernel size 3 and padding 1, producing 16, 32, and 64 output channels respectively. Each convolutional layer was followed by a ReLU activation, with max-pooling applied after the first two layers only. The output of the final convolutional layer was passed through global average pooling and then a linear layer to produce a 64-dimensional embedding. Then the base score and edge weight functions are made up of a four layer MLP with the three hidden layers containing 64 neurons each and the output of the edge weight function having 128 neurons.

\paragraph{CIFAR-10}

For CIFAR-10, a ResNet32 was used to as the DNN, outputting an embedding of length 64. Then the base score and edge weight functions are made up of a four layer MLP with the first hidden layer having 64 neurons, the second hidden layer 48 neurons and the third hidden layer 32 neurons. The output layer for the edge weight function had 64 neurons.

The ResNet was pre-trained on the CIFAR-10 data using a linear layer at the end that was removed when combined with the base score and edge weight MLPs. The ResNet was trained with a batch size of 128, 200 epochs, a learning rate of 0.1 and a weight decay of 0.00005 with SGD with momentum 0.9 and a multi step LR scheduler (stepping at 100 and 150 epcohs respectively). All parameters of the ResNet and batch normalisation layers were frozen when training end-to-end the MLPs for the base score and edge weight functions.

\subsection{Hyperparameters}

Hyperparameters for the training configuration are provided in Table~\ref{tab:training-hyp}. AdamW~\cite{adamw} was use for optimisation with the learning rate and weight decay listed in the table. All other value are left as the PyTorch~\cite{pytorch} defaults. Gradient max norm refers to the maximum gradient value (normalised) allowed during training with values greater than this clipped. Gradient clipping helps stabilise training. Hyperparameters for the loss coefficients are provided in Table~\ref{tab:loss-coeff}. Deep Arguing specific hyperparameters are provided in Table~\ref{tab:misc-hyp}.
For unbalanced datasets, the cross entropy loss function was re-weighted using, where the weight for class $c$ is $\sqrt{\frac{|D_t|}{C\cdot n_c}}$ where $D_t$ is the input dataset, $C$ is the number of classes and $n_c$ is the number of data points in class $c$.

\begin{table}
  \caption{ Training hyperparameters for the learning rate (lr), epochs, batch size, max norm of the gradients and weight decay. }
  \label{tab:training-hyp}
  \centering
  \begin{tabular}{llllll}
    \toprule
    Dataset     & lr  & epochs   & batch size & gradient max norm & weight decay  \\
    \midrule
    Adult & 0.003  & 6   & 256  & 3.0 &  0.01  \\
    Bank Marketing & 0.003  & 10  & 256  & 3.0 & 0.01 \\
    Chess & 0.007  & 1024  & 1024 & 0.1 & 0.09 \\
    Covertype & 0.0005  & 32  & 2048 & 3.5 & 0.001 \\
    Glioma & 0.003  & 32  & 64 & 3.0 & 0.0001 \\
    Higgs & 0.0005  & 32  & 2048 & 4.5 & 0.001 \\
    \midrule
    CIFAR-10 & 0.00175  & 5  & 256 & 0.5 & 0.00011 \\
    MNIST & 0.003  & 10  & 256 & 3.0 & 0.001 \\
    Fashion MNIST & 0.003  & 10  & 256 & 3.0 & 0.001 \\
    \bottomrule
  \end{tabular}
\end{table}

\begin{table}
  \caption{ Coefficients of the loss terms.  }
  \label{tab:loss-coeff}
  \centering
  \begin{tabular}{lllll}
    \toprule
    Dataset     & $\lambda_\delta$  & $\lambda_{dag}$   & $\lambda_{sp}$ & $\lambda_{sp'}$  \\
    \midrule
    Adult & 1.0  & 0.0001  & 0.0001  & 0.0001     \\
    Bank Marketing & 1.0  & 0.0001  & 0.0001 & 0.0001 \\
    Chess & 1.0   & 0.1  & 0.00001 & 0.0001 \\
    Covertype & 1.0  & 0.00001  & 0.001 & 0.001 \\
    Glioma & 1.0   & 0.0001  & 0.0001 & 0.0001 \\
    Higgs & 1.0  & 0.00001  & 0.01 & - \\
    \midrule
    CIFAR-10 & 1.0  & 0.0001  & 0.001 & - \\
    MNIST & 1.0  & 0.0001  & 0.0000 & 0.0000 \\
    Fashion MNIST &  1.0  & 0.0001  & 0.0000 & 0.0000 \\
    \bottomrule
  \end{tabular}
\end{table}

\begin{table}
  \caption{ Deep Arguing specific hyperparameters where $\alpha$ is the temperature value described in Section~\ref{sec:methodology:functions}, $I$ is the maximum number of iterations in the semantics and $t$ is the temperature of the LogSumExp operation described in Section~\ref{sec:min}.  }
  \label{tab:misc-hyp}
  \centering
  \begin{tabular}{lllll}
    \toprule
    Dataset     & $\alpha$ & no. clusters   & $I$ & $t$   \\
    \midrule
    Adult &  85  & 5   &  65  & 0.025     \\
    Bank Marketing & 125  & 5  & 65 & 0.025 \\
    Chess & 100  & 5  & 90 & 0.025 \\
    Covertype & 100  & 5 & 10 & 0.025 \\
    Glioma & 10  & 5  & 5 & 0.025 \\
    Higgs & 100  & 5  & 10 & 0.025 \\
    \midrule
    CIFAR-10 & 1000.0  & 5  & 10  & 0.025\\
    MNIST & 100  & 5  & 60 & 0.025 \\
    Fashion MNIST & 100  & 5  & 60  & 0.025\\
    \bottomrule
  \end{tabular}
\end{table}

\subsection{Compute Resources}
\label{appendix:compute-resources}

Models were trained on either a PC with an RTX 3070 GPU, AMD Ryzen 7 3700x (16 core) and 32GB of RAM or on a PC with an RTX 4070ti GPU, 13th Gen i7-13700K with 32GB of RAM. On the Deep Arguing evaluations, the combined training and evaluation time across all datasets for the 9 datasets was 4 hours. This does not include automated hyperparameter tuning time which took a total of 72 hours and manual hyperparameter tuning by one person which took 16 hours.

%% file: appendix/experiment-discussion.tex
As noted in Section~\ref{sec:experiments}, we observe that Gradual AA-CBR was unable to run on 5 out of the 6 tabular dataset
.
This is because 
it uses the full training set as the casebase, but both the original implementation and our re-implementation require the use of adjacency matrices to represent the casebase, leading to a matrix that is squared in the size of the training data. So, even with a smaller batch size, this leads to an out of memory error. 
For the one dataset, Glioma, in which this baseline was able to run, we see far superior performance with Deep Arguing. This is because Gradual AA-CBR is prone to class collapse, wherein most predictions are made for a single class. This could be attributable to a few flaws with the model. Firstly the feature extractors are small in capacity and therefore unable to effectively learn necessary feature representations. Secondly, the feature extractor emits a single value that is used for comparing cases, which may not be useful capturing the necessary dimensionality of the latent features and can lead to pairwise cycles in the graph. Finally, the model has no considerations for gradient flow, using the full dataset for the casebase, a product T-norm and no optimisations for reducing cycles, so even when increasing the model capacity, the model still fails to learn. Any one of these reasons may lead to class collapse.

%% file: appendix/full-qbaf.tex
\begin{figure}
    \centering
    \includegraphics[width=1\linewidth]{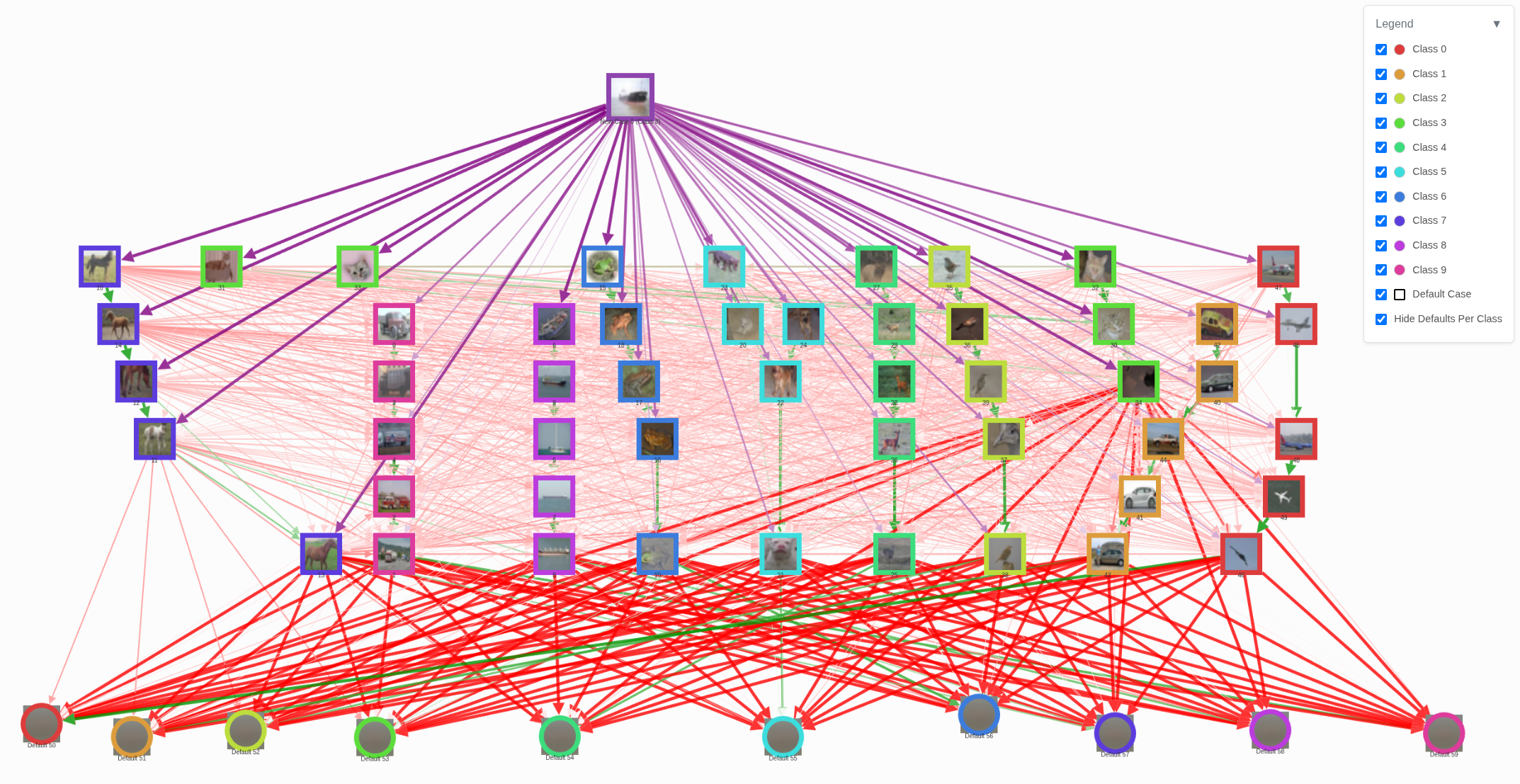}
    \caption{This figure showcases the full QBAF generated for the example in Section~\ref{sec:interpretability}. The red arrows indicate attacks, the green arrows supports and the purple arrows are irrelevance attacks. }
    \label{fig:full-qbaf}
\end{figure}

Figure~\ref{fig:full-qbaf} demonstrates the full QBAF for the example in Figure~\ref{fig:cifar10-example}.\footnote{Note that when visualising the subgraph in Figure~\ref{fig:cifar10-example}, the class colourings are different than in Figure~\ref{fig:full-qbaf}. This is purely a visualisation choice providing clearer distinction between the two selected classes, but note that the subgraph in Figure~\ref{fig:cifar10-example} is derived from the full QBAF presented here.} Whilst the full graph is complex, given the nature of many interactions within the casebase, by visualising subsets of the graph, we can successfully generate explanations that compare classes of interest to identify not just why a new case is classed as it is, but also why it is not classed differently.